\documentclass[10pt,twocolumn,letterpaper]{article}

\usepackage{iccv}
\usepackage{times}
\usepackage{epsfig}
\usepackage{graphicx}
\usepackage{amsmath}
\usepackage{amssymb}

\makeatletter
\@namedef{ver@everyshi.sty}{}
\makeatother

\usepackage[pagebackref=true,breaklinks=true,letterpaper=true,colorlinks,bookmarks=false]{hyperref}

\iccvfinalcopy 

\def\iccvPaperID{1207}

\ificcvfinal\fi

\usepackage[ruled,titlenumbered,vlined]{algorithm2e}
\usepackage{booktabs}
\usepackage{soul}
\usepackage{url}
\usepackage[utf8]{inputenc}
\usepackage{cuted}
\usepackage{amssymb}
\usepackage{xcolor}
\usepackage{multicol}
\usepackage{multirow}
\usepackage{makecell}
\usepackage{pifont}

\usepackage{wrapfig}
\usepackage{mathtools}
\usepackage{colortbl}
\usepackage{float}
\usepackage[position=bottom]{subfig}
\usepackage{tikz}
\usepackage{pgfplots}
\usepackage{pgfplotstable}
\usepackage{cleveref}
\crefname{section}{Sec.}{Secs.}
\Crefname{section}{Section}{Sections}
\Crefname{table}{Table}{Tables}
\crefname{table}{Tab.}{Tabs.}
\Crefname{equation}{Equation}{Equations}
\crefname{equation}{Eq.}{Eqs.}
\Crefname{algorithm}{Algorithm}{Algorithms}
\crefname{algorithm}{Alg.}{Algs.}
\Crefname{algocf}{Algorithm}{Algorithms}
\crefname{algocf}{Alg.}{Algs.}
\Crefname{figure}{Figure}{Figures}
\crefname{figure}{Fig.}{Figs.}
\Crefname{appendix}{Appendix}{Appendixes}
\crefname{appendix}{Appendix}{Appendixes.}

\usepackage{tabularx}
\newcolumntype{Y}{>{\centering\arraybackslash}X}

\usepackage[english]{babel}
\usepackage{amsthm}
\usepackage{subfig}
\definecolor{Gray}{gray}{0.85}

\definecolor{lightgreen}{RGB}{86,188,80}

\allowdisplaybreaks

\renewcommand{\eg}{\emph{e.g.}}
\renewcommand{\ie}{i.\,e.}
\newcommand{\mcref}[1]{Eq.~\ref{#1}}

\usepackage{colortbl}
\colorlet{LGray}{gray!10}

\newcommand\blfootnote[1]{
  \begingroup
  \renewcommand\thefootnote{}\footnote{#1}
  \addtocounter{footnote}{-1}
  \endgroup
}

\allowdisplaybreaks

\begin{document}

\title{Event Camera Data Pre-training}

\author{
Yan Yang$^{1}$ \quad Liyuan Pan$^{2~\dagger}$ \quad Liu Liu$^{3}$ \\
$^1$BDSI, ANU \quad  $^2$BITSZ \& School of CSAT, BIT \quad  $^3$Cyberverse Dept., Huawei \\
{\tt \small Yan.Yang@anu.edu.au \quad liyuan.pan@bit.edu.cn \quad liuliu33@huawei.com}
}

\maketitle
\blfootnote{$^{\dagger}$ Corresponding author.}
\ificcvfinal\thispagestyle{empty}\fi

\begin{abstract}
This paper proposes a pre-trained neural network for handling event camera data. Our model is a self-supervised learning framework, and uses paired event camera data and natural RGB images for training. 

Our method contains three modules connected in a sequence: i) a family of event data augmentations, generating meaningful event images for self-supervised training; ii) a conditional masking strategy to  sample informative event patches from event images, encouraging our model to capture the spatial layout of a scene and accelerating training; iii) a contrastive learning approach, enforcing the similarity of embeddings between matching event images, and between paired event and RGB images. An embedding projection loss is proposed to avoid the model collapse when enforcing the event image embedding similarities. A probability distribution alignment loss is proposed to encourage the event image to be consistent with its paired RGB image in the feature space.

Transfer learning performance on downstream tasks shows the superiority of our method over state-of-the-art methods. For example, we achieve top-1 accuracy at 64.83\% on the N-ImageNet dataset. 
\end{abstract}

\section{Introduction}
\label{sec:intro}
An event camera asynchronously captures the time, location, and polarity of pixel-wise changes in brightness as a sequence of events. Event cameras are widely used in many applications, \eg, recognition \cite{nimagnet}, detection \cite{od1,od3},  segmentation \cite{ddd17}, optical flow estimation \cite{mvsec}, and SLAM \cite{slam}. Compared with conventional RGB cameras which record 
all pixel intensities at a fixed frame rate, event cameras enjoy a high dynamic range and temporal resolution, and are robust to lighting changes and motion blur \cite{nimagnet,ess,ed1}.

\begin{figure}[!t]
    \centering
        \includegraphics[width=.8\linewidth]{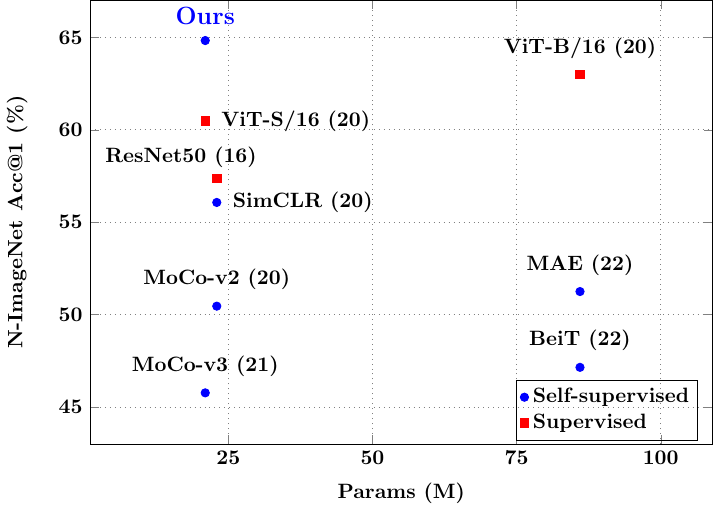}
    \vspace{-1em}
    \caption{\it \small
    Comparison of our methods and state-of-the-art methods on N-ImageNet dataset \cite{nimagnet}. The \textcolor{blue}{Blue} cycles and \textcolor{red}{red} squares separately denote the self-supervised and supervised pre-training methods. We show top-1 accuracy (\%), \ie, acc@1, with respect to the number of model parameters (M). We include the publication year of each method in the brackets beside the method names. Best viewed in color on the screen.
    }
    \label{fig:overall_comaprision}
\end{figure}

This paper studies the problem of event camera data pre-training. Our model is pre-trained in a self-supervised manner, only using paired event data and RGB images for training. One can simply transfer our pre-trained model for diverse downstream tasks.

In self-supervised learning (SSL), significant progress has been made in pre-training with RGB images \cite{mae,beit,mocov3}. However, it is non-trivial to replicate the success on event camera data, as there is a domain gap between RGB images and event data. An RGB image records all pixel intensities of a scene and is spatially dense, while the event data only records scene changes and is spatially sparse.

For network training in the SSL framework, image augmentations (\eg, Gaussian Blur, ColorJitter, RandomResizedCrop) are one of the most important parts. The sparse event camera data can be commonly represented as an event image \cite{nimagnet}. One may directly and wrongly perform these augmentations on event images, \eg, blurring a binary event image (0/1 valued pixels) generates a meaningless event image. In contrast, we study how to perform event data augmentations before converting to an event image.

We formulate our learning problem as a contrastive learning task. Taking event images as inputs, one may directly perform a random masking strategy to sample a fixed number of event patches for encouraging the model to capture the spatial layout and accelerating training. However, an event image is spatially sparse, and random masking would generate non-informative patches, leading to training instability. To mitigate this problem, we propose a conditional masking strategy to sample informative patches.

With event patches, we are able to learn discriminative event embeddings, \ie, pulling together embeddings from similar event images while pushing away embeddings from dissimilar ones. Surprisingly, we find that simply performing metric learning in the event embedding space leads to model collapse, producing over-similar embeddings. The reason comes from the spatial sparsity of event images. To solve this problem,  we find that embeddings from paired RGB images can be used as a regularizer, and we propose an embedding projection loss to solve the collapse.

With paired event data and RGB images, we also aim to pull together embeddings from matched pairs. This is motivated by the fact that many well pre-trained RGB networks are available, and an event image is less informative than its paired RGB image. Therefore, the RGB network serves as a teacher for our event network, and we propose a probability distribution alignment loss for the learning.

Our contributions are summarized as follows:
\vspace{-.5em}
\begin{itemize}
    \setlength{\itemsep}{-.2em}
    \item A self-supervised framework for event camera data pre-training. The pre-trained model can be transferred to diverse downstream tasks;
    \item A family of event data augmentations, generating meaningful event images;
    \item A conditional masking  strategy, sampling informative event patches for network training;
    \item An embedding projection loss, using paired RGB embeddings to regularize event embeddings to avoid model collapse;
    \item A probability distribution alignment loss for aligning embeddings from the paired event and RGB images.
    \item We achieve state-of-the-art performance in standard event benchmark datasets (\eg, Fig. \ref{fig:overall_comaprision}). 
\end{itemize}

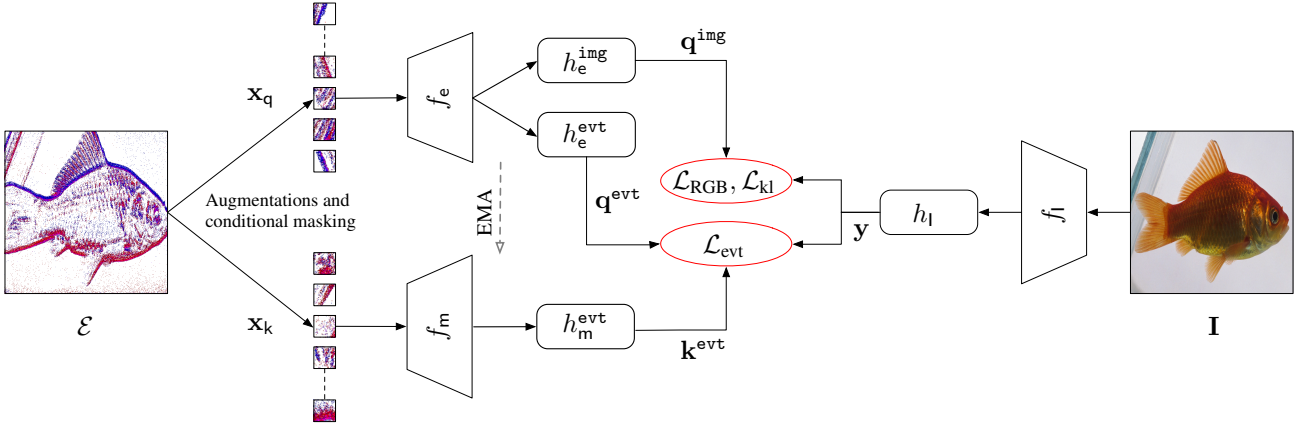
\begin{figure*}[!t]
    \centering
        \begin{tikzpicture}
        \tikzstyle{every node}=[font=\scriptsize]
        \centering
        \node[anchor=south west,inner sep=0] (image) at (0,0) {\includegraphics[width=\linewidth]{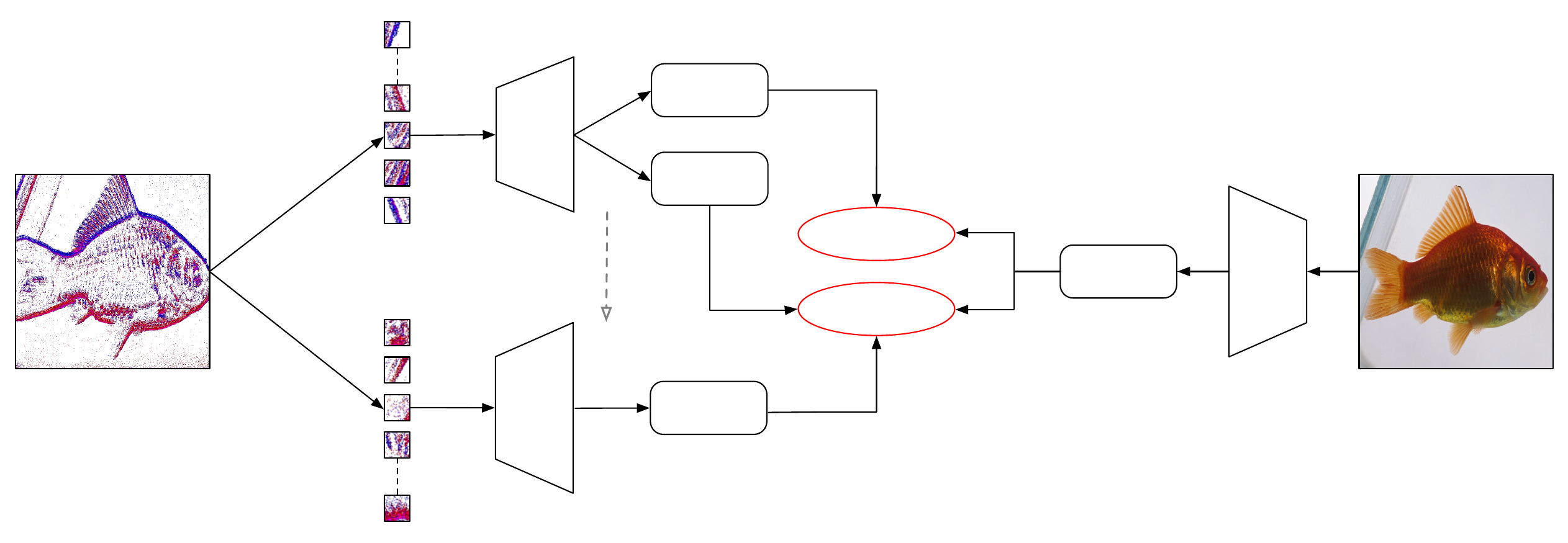}};
        \begin{scope}[x={(image.south east)},y={(image.north west)}]
        \draw (0.0715, 0.25) node {\normalsize $\mathcal{E}$};
        \draw (0.22,0.5) node {\makecell[l]{Augmentations and\\conditional masking}};
        \draw (0.205, 0.25) node {\normalsize $\mathbf{x}_{\mathsf{k}}$};
        \draw (0.205, 0.75) node {\normalsize $\mathbf{x}_{\mathsf{q}}$};
        \draw (0.340, 0.25) node [rotate=90] {\normalsize $f_{\mathsf{m}}$};
        \draw (0.340, 0.75) node [rotate=90] {\normalsize $f_{\mathsf{e}}$};
        \draw (0.452, 0.251) node {\normalsize $h^{\tt{evt}}_{\mathsf{m}}$};
        \draw (0.452, 0.669) node {\normalsize $h^{\tt{evt}}_{\mathsf{e}}$};
        \draw (0.452, 0.833) node {\normalsize $h^{\tt{img}}_{\mathsf{e}}$};
        \draw (0.709, 0.495) node {\normalsize $h_{\mathsf{I}}$};
        \draw (0.808, 0.495) node [rotate=90]{\normalsize $f_{\mathsf{I}}$};
        \draw (0.93, 0.25) node {\normalsize $\mathbf{I}$};
        \draw (0.557, 0.565) node {\normalsize $\mathcal{L}_{\text{RGB}},\mathcal{L}_{\text{kl}}$};
        \draw (0.557, 0.425) node {\normalsize 
        \makecell{
        $\mathcal{L}_{\text{evt}}$} 
        };
        \draw (0.375, 0.5) node [rotate=90] {\footnotesize EMA};
        \draw (0.542, 0.20) node {\normalsize ${\mathbf{k}}^{\tt{evt}}$};
        \draw (0.477, 0.525) node {\normalsize ${\mathbf{q}}^{\tt{evt}}$};
        \draw (0.542, 0.875) node {\normalsize ${\mathbf{q}}^{\tt{img}}$};
        \draw (0.662, 0.455) node {\normalsize ${\mathbf{y}}$};
        \end{scope}
    \end{tikzpicture}
    \vspace{-2.5em}
    \caption{\it  \small
    The overall architecture. For pre-training, our method takes event data $\mathcal{E}$ and its paired natural RGB image $\mathbf{I}$ as inputs, and outputs a pre-trained network $f_\mathsf{e}$. Given $\mathcal{E}$ (its abstract representation is used for visualization purposes), we first consecutively perform data augmentations, event image generation, and conditional masking to obtain two patch sets $(\mathbf{x}_{\mathsf{q}}, \mathbf{x}_{\mathsf{k}})$. 
    Second, $f_\mathsf{e}$ extracts features from event patch set $\mathbf{x}_{\mathsf{q}}$, and $h^{\tt{img}}_{\mathsf{e}}$ and $h^{\tt{evt}}_{\mathsf{e}}$ separately project features from $f_\mathsf{e}$  to latent embeddings ${\mathbf{q}}^{\tt{img}}$ and  ${\mathbf{q}}^{\tt{evt}}$. $f_{\mathsf{m}}$ and $h^{\tt{evt}}_{\mathsf{m}}$ are the momentum of $f_{\mathsf{e}}$ and $h^{\tt{evt}}_{\mathsf{e}}$, and are updated by the exponential moving average (EMA). The momentum network takes patch set $\mathbf{x}_{\mathsf{k}}$ as input and generates an embedding  ${\mathbf{k}}^{\tt{evt}}$.
    At the same time, the natural RGB image $\mathbf{I}$ is embeded into ${\mathbf{y}} = f_{\mathsf{I}}(h_{\mathsf{I}}(\mathbf{I}))$. Finally, we perform event discrimination, and event and natural RGB image discrimination to train our model. We optimize the network by $\mathcal{L}_\text{evt}$ (\mcref{eq:lx}), $\mathcal{L}_\text{RGB}$ (\mcref{eq:ly}), and $\mathcal{L}_\text{kl}$ (\mcref{eq:kl}). $\mathcal{L}_{\text{evt}}$ is an event embedding projection loss aiming to pull together paired event embeddings ${\mathbf{q}}^{\tt{evt}}$ and ${\mathbf{k}}^{\tt{evt}}$, for event discrimination. $\mathcal{L}_{\text{RGB}}$ aims to pull together paired event and RGB embeddings ${\mathbf{q}}^{\tt{evt}}$ and ${\mathbf{y}}$, for  event and natural RGB image discrimination.  $\mathcal{L}_{\text{kl}}$ aims to drive $f_{\mathsf{e}}$ learning discriminative event embeddings, towards well-structured embedding space of natural RGB images. Best viewed in color. 
    }
    \label{fig:arch}
\end{figure*}

\section{Related Work}
The SSL frameworks can be generally divided into two categories: contrastive learning and masked modeling. We briefly review their recent achievements, and then introduce event datasets for diverse computer vision tasks.

\vspace{-4mm}
\paragraph{Contrastive learning.} This approach generally assumes augmentation invariance of images \cite{simclr,mocov1}. Two or more views of each image are generated for instance discrimination that enforces embedding similarity and dissimilarity among the views \cite{simclr,mocov1,mocov2,mocov3,instancediscrmination}.  Only enforcing the embedding similarity is also possible and has been studied in \cite{simsiam,byol}. In addition to model design and optimization objectives,  contrastive learning approaches usually rely on strong augmentations over images to boost model performance \cite{simclr,mocov1,mocov2,mocov3,swav,simsiam}. Under certain tasks, contrastive learning has shown better performance than supervised pre-training \cite{vit,mocov3}. However, one notable drawback of contrastive learning is suffering from model collapse and training instability. Diverse methods including asymmetric network designs (\eg, maintaining a momentum network) \cite{simclr,mocov1}, partial weight freezing \cite{mocov3}, and group-based discrimination \cite{swav, dino} are introduced to avoid the model collapse and instability issue. 

\vspace{-4mm}
\paragraph{Masked modeling.} Reconstructing masked inputs from the (\ie, unmasked) visible ones is a popular self-supervised learning objective motivated by the idea of auto-encoding. The pioneer works can be specified to the natural language processing domain, \eg,  Bert \cite{bert} and GPT \cite{gpt}. Recently, masked modeling has been formulated in the image domain, where the objective is defined in a similar vein, and the masking of images is done pixel-wisely or patch-wisely \cite{igpt,mae,vit,beit,ibot}. Some works \cite{beit,ibot} turn the masked modeling into a classification problem by predicting discrete indices assigned to the masked patches by a tokenizer, \eg, pre-trained discrete VAE \cite{dvae,vqgan} or self-distilled network \cite{ibot,dino}. One could also target to directly regress the pixel intensity of masked patches \cite{igpt,mae,vit}.

\vspace{-4mm}
\paragraph{Event datasets.} The event camera is a novel sensor that asynchronously captures the time, location, and polarity (\ie, direction) of per-pixel brightness change as a sequence of events. With growing interest in event-based computer vision tasks, researchers have collected a wide range of datasets for object recognition \cite{nimagnet,ncars,ncaltech,CIFAR-10-DVS}, semantic segmentation \cite{ddd17}, optical flow estimations \cite{mvsec}, and so forth \cite{rc1,rc2,rc3}. To leverage existing computer vision algorithms, \eg, CNN and ViT, the majority of event-based vision frameworks convert event data into image/video-liked grid representations, where the conversion is done either learnable \cite{eventtoimage} or by directly using the position and time of each event \cite{nimagnet}. This paper leverages the event image representation to study the event-based SSL algorithm that benefits diverse event-based downstream tasks.

\section{Method}
\label{sec:method}

We start with a brief overview of background knowledge, and then present our self-supervised learning framework, in this section. Our network is trained end-to-end, and the overall architecture is shown in \cref{fig:arch}. 

\vspace{-4mm}
\paragraph{Preliminary.} 
Contrastive learning aims to learn an embedding space, where similar image pairs stay close to each other while dissimilar ones are far apart.
Specifically, images are embedded into vectors to collect a query set $\{\mathbf{q}\}$ and a key set $\{\mathbf{k}\}$. For each query $\mathbf{q}$, we have a matching key $\mathbf{k}_{+}$ and non-matching keys $\{\mathbf{k}_{-}\}$. Usually, $\mathbf{q}$ and $\mathbf{k}_{+}$ are generated from views of the same instance, while $\mathbf{q}$ and $\{\mathbf{k}_{-}\}$ are generated from views of different instances. Contrastive learning aims to pull together embeddings $\mathbf{q}$ and $\mathbf{k}_{+}$, and pushes away embeddings $\mathbf{q}$ and $\{\mathbf{k}_{-}\}$. In this paper, we use the InfoNCE loss \cite{infonce},
\begin{align}\label{Eq::infonce}
\mathcal{L}_\text{nce}(\mathbf{q},\{\mathbf{k}\}) = - \log \frac{\exp(\mathbf{q} \cdot \mathbf{k}_{+} / \tau )}{\exp(\mathbf{q} \cdot \mathbf{k}_{+} / \tau ) + \sum\limits_{\mathbf{k}_{-}} \exp(\mathbf{q} \cdot \mathbf{k}_{-} / \tau )} \ ,
\end{align}
where $\mathbf{q}$ and $\mathbf{k}$ are $L_2$ normalized to a metric space, and the similarity between them is then measured by the cosine similarity using dot-product $(\cdot)$. $\tau$ is a temperature hyper-parameter \cite{mocov3}.

\vspace{-4mm}
\paragraph{Overall architecture.} 
Given an event data $\mathcal{E} = {(u_{i},t_{i},p_{i})}^{\mathcal{N}}_{i=1}$ and a paired natural RGB image $\mathbf{I}$, where $u_{i}$, $t_{i}$, and $p_{i}$ separately denotes spatial location, time, and polarity of each event, and $\mathcal{N}$ is the length of the event data.

We aim to pre-train a neural network $f_\mathsf{e}$, such that $f_\mathsf{e}$ can generate discriminative features for benefiting diverse event-based downstream tasks. Our method is self-supervised and has three components: i) event image patch generation. Given input $\mathcal{E}$, it generates matching patches $(\mathbf{x}_{\mathsf{q}}, \mathbf{x}_{\mathsf{k}})$ on $\mathcal{E}$;   ii) event discrimination. It aims to pull together embeddings of $(\mathbf{x}_{\mathsf{q}}, \mathbf{x}_{\mathsf{k}})$; iii) event and RGB image discrimination. It aims to pull together embeddings of $\mathbf{x}_{\mathsf{q}}$ and $\mathbf{I}$. Details of the above three components are given in the following paragraphs.

\vspace{-4mm}
\paragraph{Event image patch generation.}
To convert $\mathcal{E}$ into two matching patches $(\mathbf{x}_\mathsf{q}, \mathbf{x}_\mathsf{k})$, we consecutively apply our data augmentations,  event image generation, and conditional masking strategy. We perform event data augmentations before converting them to event images. Please refer to the supplementary material for details, \ie, how to perform RandomResizedCrop, GaussianBlur, and ColorJitter for $\mathcal{E}$. With augmented $\mathcal{E}$,
we first generate an event image by applying the event histogram algorithm \cite{eventhistorgram}, and then use our conditional masking strategy to obtain patches $\mathbf{x}_\mathsf{q}$ and patches $\mathbf{x}_\mathsf{k}$. 

Given an event image, considering its sparsity, using a random masking strategy to sample patches is prone to generate meaningless/non-informative patches. Therefore, a conditional masking strategy is proposed to sample patches. Let $\{\mathbf{p}_{i}\}^{\mathcal{P}}_{i=1}$ be a patch set of an event image, $\mathbf{p}_{i}$ is the $i$-th patch, and $\mathcal{P}$ is the cardinality of the set. After vectorizing $\mathbf{p}_{i}$, we calculate the information quantity $\mathsf{d}_{i}$ of each patch,  
\begin{equation}
\mathsf{d}_{i} = \lvert \mathbf{p}_{i} \rvert \cdot \mathbf{1}, \quad \forall i \in [1,\cdots, \mathcal{P}],
\end{equation}
where $\mathbf{1}$ denotes a vector of ones. Collecting $\mathcal{P}$ information quantities and $L_1$ normalizing them, we obtain a probability distribution. A patch probability describes how likely it contains meaningful information. 
We randomly sample a fixed number ($\ll \mathcal{P}$) of patches according to the probability distribution, resulting in $\mathbf{x}_\mathsf{q}$. Then, the same process is performed to generate $\mathbf{x}_\mathsf{k}$. 

\vspace{-4mm}
\paragraph{Event discrimination.}
With patches $\mathbf{x}_\mathsf{q}$ and patches $\mathbf{x}_\mathsf{k}$, we show how to pull together embeddings of them. $\mathbf{x}_\mathsf{q}$ is fed to network $f_\mathsf{e}$ to extract features, and features from $f_\mathsf{e}$ are fed to a projection head $h^{\tt{evt}}_{\mathsf{e}}$ to extract an embedding ${\mathbf{q}}^{\tt{evt}}$, ${\mathbf{q}}^{\tt{evt}} = h^{\tt{evt}}_{\mathsf{e}} (f_{\mathsf{e}}(\mathbf{x}_\mathsf{q}))$. For self-supervised training, $\mathbf{x}_\mathsf{k}$ is fed to $f_\mathsf{m}$ and $h^{\tt{evt}}_{\mathsf{m}}$ to extract an embedding ${\mathbf{k}}^{\tt{evt}}$, ${\mathbf{k}}^{\tt{evt}} = h^{\tt{evt}}_{\mathsf{m}}( f_{\mathsf{m}}(\mathbf{x}_\mathsf{k}))$, where  $f_\mathsf{m}$ and $h^{\tt{evt}}_{\mathsf{m}}$ are the momentum \cite{mocov1} of $f_\mathsf{e}$ and $h^{\tt{evt}}_{\mathsf{e}}$, respectively.

To enforce the similarity between embeddings ${\mathbf{q}}^{\tt{evt}}$ and ${\mathbf{k}}^{\tt{evt}}$, one may directly optimize the InfoNCE loss $\mathcal{L}_\text{nce}({\mathbf{q}}^{\tt{evt}},  \{ {\mathbf{k}}^{\tt{evt}} \})$.  However, we find that optimized embeddings collapse, \ie, they are over-similar. The reason would be the sparsity of event images, and the sparsity decreases the discriminativeness of event embeddings. 

To solve this collapse problem, interestingly, we find that the embedding ${\mathbf{y}} = h_{\mathsf{I}}(f_{\mathsf{I}}(\mathbf{I}))$ of the paired natural RGB image $\mathbf{I}$ is a basis vector and provides good regularization. $f_{\mathsf{I}}$ is an image feature extraction network, and $h_{\mathsf{I}}$ projects features to an embedding.  
We have the event embedding projection loss,
\begin{align}
    \mathcal{L}_\text{evt} &= \mathcal{L}_\text{nce}\big(\zeta({\mathbf{q}}^{\tt{evt}}, {\mathbf{y}}),  \{ \zeta({\mathbf{k}}^{\tt{evt}},{\mathbf{y}}) \} \big) \ , \\
    \zeta(\mathbf{v}_{1},\mathbf{v}_{2}) &= \mathbf{v}_{1} \cdot \mathbf{v}_{2} ~ \frac{\mathbf{v}_{2}}{\lVert \mathbf{v}_{2} \rVert} , 
    \label{eq:lx}
\end{align}
where $\zeta(\mathbf{v}_{1},\mathbf{v}_{2})$ is the projection function. Here, $\zeta({\mathbf{q}}^{\tt{evt}}, {\mathbf{y}})$ and $\zeta({\mathbf{k}}^{\tt{evt}},{\mathbf{y}})$ separately projects event embeddings ${\mathbf{q}}^{\tt{evt}}$ and ${\mathbf{k}}^{\tt{evt}}$ to embedding ${\mathbf{y}}$. 
We do not perform $L_2$ normalization on  $\zeta({\mathbf{q}}^{\tt{evt}}, {\mathbf{y}})$ and $\zeta({\mathbf{k}}^{\tt{evt}},{\mathbf{y}})$ for calculating $\mathcal{L}_\text{evt}$.

\vspace{-4mm}
\paragraph{Event and RGB image discrimination.} Considering the sparsity of the event image, a single event image is less informative than an RGB image, possessing difficulty for self-supervised event network training. In contrast, many well-trained RGB networks are available. We aim to teach our event network $f_\text{e}$, using well pre-trained RGB network $f_{\mathsf{I}}$. We pull together embeddings of paired event and RGB images, $\mathbf{x}_{\mathsf{q}}$ and $\mathbf{I}$. Features from $f_\text{e}$ are fed to a projection head $h^{\tt{img}}_{\mathsf{e}}$ to extract an event image embedding ${\mathbf{q}}^{\tt{img}}$.
Given embeddings ${\mathbf{q}}^{\tt{img}}$ and ${\mathbf{y}}$, we enforce their similarity by optimizing the InfoNCE loss,
\begin{align}
    \mathcal{L}_\text{RGB} = \mathcal{L}_\text{nce} ({\mathbf{q}}^{\tt{img}}, \{{\mathbf{y}}\}) .
    \label{eq:ly}
\end{align}
To better align event and RGB embedding spaces, we first separately fit two probability distributions in the event and RGB embedding spaces, {and then use Kullback–Leibler divergence to minimize the distribution mismatch.}

Specifically, given a batch of event embeddings $\{{\mathbf{q}}^{\tt{img}}\}$, we first compute the pairwise embedding similarity and then fit an exponential kernel to the similarities to compute probability scores. The probability score of the $(i,j)$-th pair is given by,
\begin{align}
{{s}}^{\mathbf{q}}_{i,j} = \frac{\exp({\mathbf{k}}_{i} \cdot {\mathbf{k}}_{j} / \tau )}{ \sum\limits_{j} \exp(\mathbf{k}_{i} \cdot \mathbf{k}_{j} / \tau )} \ ,
\end{align}
where $\mathbf{k}_{i}$ and $\mathbf{k}_{j}$ are the $i$-th and $j$-th embedding of the batch $\{{\mathbf{q}}^{\tt{img}}\}$. $\tau$ is the same hyperparameter in Eq.~\eqref{Eq::infonce}. 
The probability score of ${\mathbf{y}}$ is obtained in the same way and is denoted as ${{s}}^{\mathbf{y}}_{i,j}$.

Our probability distribution alignment loss is given by,
\begin{equation}
    \mathcal{L}_\text{kl} = \sum_{i}\sum_{j}{s^\mathbf{q}_{i,j}}\cdot \log\left( \frac{s^\mathbf{q}_{i,j}}{s^\mathbf{y}_{i,j}} \right) \label{eq:kl}
\end{equation}

\vspace{-4mm}
\paragraph{Losses.} 
Our network is trained end-to-end, and the total loss is 
\begin{equation}
    \mathcal{L}_{total} = \mathcal{L}_\text{evt} +  \mathcal{L}_\text{RGB} + \lambda_{1} \mathcal{L}_\text{kl},
\end{equation}
where $\lambda_{1}$ is a hyper-parameter for balancing the losses.

\begin{table*}[!t]
    \centering
    \caption{\it \small  
    Comparison of object recognition accuracies on the N-ImageNet dataset \cite{nimagnet}.
    We show the top-1 and top-5 accuracies (\ie, acc@1 and acc@5 (\%)) of state-of-the-art methods.  
    }
    \vspace{-1em}
    \small
    \begin{tabularx}{\linewidth}{XYYYYY}
        \toprule
        \multicolumn{1}{c}{\multirow{2}{*}{Method}} & \multirow{2}{*}{Architecture} & \multirow{2}{*}{Parameters} & \multirow{2}{*}{Pre-training Epoch} & \multicolumn{2}{c}{Fine-tuning} \\
        \cmidrule{5-6}  
        & & & & acc@1  & acc@5 \\
        \midrule
        \multicolumn{6}{l}{\textit{The best performance in the literature. }}   \\ 
        EST~\cite{est} & - & 21M & - & 48.93 & - \\
        \midrule
        \multicolumn{6}{l}{
        \textit{Training from scratch, \ie, random weight initialization.}
        } \\
        ViT \cite{vit} & ViT-S/16 & 21M  &  -  &  46.70 & 69.89\\
        ViT \cite{vit} & ViT-B/16 & 86M  &  -  & 51.23 & 74.50\\
        ResNet \cite{resnet} & ResNet50 & 23M & - & 50.07 & 74.83 \\ 
        \midrule
        \multicolumn{6}{l}{\textit{Transfer learning of supervised pre-training methods, \ie, initial weights learned in a supervised manner.
        }} \\
        ViT \cite{vit} & ViT-S/16 & 21M & 300 & 60.48 & 83.02\\
        ViT \cite{vit} & ViT-B/16 & 86M & 300 & 62.98 & 84.75\\
        ResNet \cite{resnet} & ResNet50 & 23M  & 90 & 57.37 & 80.93\\
        \midrule 
        \multicolumn{6}{l}{\textit{Transfer learning of self-supervised pre-training methods, \ie, initial weights learned in a self-supervised manner. }} \\
        SimCLR \cite{simclr}  & ResNet50 & 23M  & 100 &  56.07 & 80.49\\
        MoCo-v2 \cite{mocov2} & ResNet50 & 23M  & 200 &  50.46 & 75.67\\
        MoCo-v3 \cite{mocov3} & ViT-S/16 & 21M  & 300 &  45.77 & 68.89\\
        BeiT \cite{beit} & ViT-B/16 & 86M & 800 & 47.15 & 69.27\\
        iBoT \cite{ibot} & ViT-S/16 & 21M & 800 & 19.55 & 38.72 \\
        MAE  \cite{mae} & ViT-B/16 & 86M  & 800 & 51.25 & 72.64\\
        \rowcolor{LGray}
        Ours & ResNet50 & 23M & 300 & {59.80} & {82.04} \\
        \rowcolor{LGray}
        Ours & ViT-S/16 & 21M & 300 & \textbf{64.83} & \textbf{86.30} \\
        \bottomrule
    \end{tabularx}
    \label{tab:obj}
\end{table*}

\section{Experiments}
\subsection{Experimental Setup}

\paragraph{Pre-training dataset.} We use the N-ImageNet \cite{nimagnet} and ImageNet-1K \cite{imagenet} datasets for pre-training. The N-ImageNet dataset is built from the ImageNet-1K dataset, where a moving event camera observes natural RGB images displayed by a monitor. Similar to the ImageNet-1K, it contains $1,781,167$ samples of event data, covering $1,000$ object classes. All event samples are recorded in $480 \times 640$ resolution. We resize them to $224 \times 224$ resolution, and use the official training set for pre-training. 

\vspace{-4mm}
\paragraph{Implementation.} 
We explore two backbones ViT-S/16 and ResNet50 for our method, and separately report our pre-training performance. We use the backbone for $f_{\mathsf{e}}$ and $f_{\mathsf{m}}$, and the same projection head as MoCo-v3 for $h^{\tt{evt}}_{\mathsf{e}}$, $h^{\tt{evt}}_{\mathsf{m}}$, and $h^{\tt{img}}_{\mathsf{e}}$. 
We use SSL pre-trained ViT-B/32 for the RGB image backbone $f_{\mathsf{I}}$, and set $h_{\mathsf{I}}$ to a single linear layer. The hyper-parameters $\lambda_{1}$ is set to $2$. Please refer to the supplement material for optimization schemes, ablation of $f_{\mathsf{e}}$, $f_{\mathsf{I}}$ and $h_{\mathsf{I}}$. 
Our method is implemented in Pytorch \cite{pytorch}. All codes and pre-trained models will be released.

\vspace{-4mm}
\paragraph{Transfer learning tasks.} We evaluate our method and state-of-the-art methods on three downstream tasks:  object recognition (\cref{sec::object_recognition}), optical flow estimation (\cref{sec::flow_estimation}), and  semantic segmentation (\cref{sec::semantic_segmentation}).

\vspace{-4mm}
\paragraph{Baselines.}
Our method is compared with four groups of methods: 
i) Previous best. We compare with state-of-the-art methods for each task;
ii) Training from scratch. We train state-of-the-art methods with random weight initialization; 
iii) Transfer learning of supervised pre-training. The initial weights of state-of-the-art methods are obtained in a supervised manner using the  ImageNet-1K dataset;
iv) Transfer learning of self-supervised pre-training. The initial weights of state-of-the-art methods are obtained in a self-supervised manner using the  ImageNet-1K dataset.

\begin{figure}[!t]
    \centering
        \begin{tikzpicture}
        \begin{axis}[
            xlabel={Epoch},
            ylabel={Top-1 Accuracy (\%)},
            compat=newest,
            xmin=50, xmax=300,
            ymin=48, ymax=62,
            xtick={50,100, 150, 200, 250, 300},
            xticklabels={50,100, 150, 200, 250, 300},
            ytick={50,55,60},
            yticklabels={50,55,60},
            xticklabel style = {font=\footnotesize},
            yticklabel style = {font=\footnotesize},
            scaled x ticks = false,
            xmajorgrids,
            ymajorgrids,
            grid style={dotted, gray},
            scale only axis=true,
            width=.8\linewidth,
            height=.3\linewidth,
            legend style={at={(0.95,0.35)}}
            ]
            \addplot[blue, line width=1pt, mark=x] coordinates
            {
            (50, 50.43)
            (100,53.51)
            (150,55.33)
            (200,57.15)
            (250,59.21)
            (300,59.90)
            };
            \legend{Linear probing}
        \end{axis}
        \end{tikzpicture}
        \vspace{-1em}
        \caption{\it \small
        The linear probing accuracy of our method with respect to the number of training epochs.}
        \label{fig:linear_probing}
\end{figure}
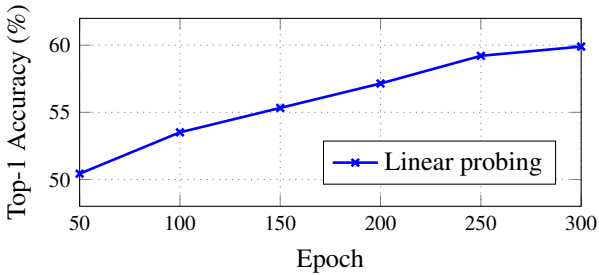

\subsection{Object Recognition}
\label{sec::object_recognition}
We first show our object recognition performance on the large-scale N-ImageNet \cite{nimagnet} dataset and then report our performance on three small-scale datasets, N-Cars \cite{ncars}, N-Caltech101 \cite{ncaltech}, and CIFAR-10-DVS \cite{CIFAR-10-DVS}. 

\vspace{-4mm}
\paragraph{Results on the large-scale N-ImageNet dataset.} 
The comparisons are given in \cref{tab:obj}. It shows that fine-tuning our pre-trained model with a ViT-S/16 backbone achieves a top-1 accuracy at 64.83\%, outperforming all other methods. {Additionally, we examine the linear probing performance of this pre-trained model, and it achieves a top-1 accuracy at 59.90\%, outperforming methods in the self-supervised group. The linear probing accuracies of our method with respect to the number of training epochs are given in \cref{fig:linear_probing}.} 

For methods (except ours) in the self-supervised group, we find that they overfit easily 
(even achieving a near-perfect top-1 training accuracy) when fine-tuning on the N-ImageNet dataset, though we have tried our best to use diverse regularization techniques. This further demonstrates
the value of this paper -- a self-supervised learning framework for event camera data pre-training.

\newcolumntype{C}[1]{>{\centering\arraybackslash}p{#1}}
\begin{table*}[!t]
    \centering
    \caption{\it  \small
    Comparison of object recognition accuracies on the N-Cars \cite{ncars}, N-Caltech101  \cite{ncaltech}, and CIFAR-10-DVS \cite{CIFAR-10-DVS} datasets. We show the top-1 accuracy for clarity.
    }
    \vspace{-1em}
    \small
    \begin{tabularx}{\linewidth}{XYYYY}
        \toprule
         \multicolumn{1}{c}{Method} & Architecture & N-Cars & N-Caltech101 & CIFAR-10-DVS \\
         \midrule
        \multicolumn{5}{l}{\textit{The best performance  in the literature. }}   \\ 
        N-ImageNet~\cite{nimagnet} & - & 94.73 & 86.81 & 73.72\\
        \midrule
        \multicolumn{5}{l}{\textit{\makecell[l]{Training from scratch, \ie, random weight initialization.}}} \\
        ViT \cite{vit} & ViT-S/16 & 89.14 & 55.63 & 52.45\\
        ViT \cite{vit} & ViT-B/16 & 93.09 & 67.11 & 55.15 \\
        ResNet \cite{resnet} & ResNet50 &  91.20 &  62.69 & 56.65  \\ 
        \midrule
        \multicolumn{5}{l}{\textit{\makecell[l]{Transfer learning of supervised pre-training methods, \ie, initial weights learned in a supervised manner.}}} \\
        ViT \cite{vit} & ViT-S/16 & 96.76 & 85.02 & 76.10\\
        ViT \cite{vit} & ViT-B/16 & 97.56 & 86.45 & 77.45\\
        ResNet \cite{resnet} & ResNet50 & 97.61 & 86.51 & 73.40\\
        \midrule 
        \multicolumn{5}{l}{\textit{\makecell[l]{Transfer learning of self-supervised pre-training methods, \ie, initial weights learned in a self-supervised manner.}}} \\
        SimCLR \cite{simclr} & ResNet50 & 97.10 & 86.57& 75.15\\
        MoCo-v2 \cite{mocov2} & ResNet50 & 96.64 &84.16 & 74.65\\
        MoCo-v3 \cite{mocov3} & ViT-S/16 & 95.33  & 76.59 & 68.40\\ 
        BeiT \cite{beit} & ViT-B/16 & 90.61 & 53.10 & 53.15\\  
        iBoT \cite{ibot} & ViT-S/16 & 92.30 & 47.36 & 56.10\\ 
        MAE \cite{mae} & ViT-B/16 & 95.34 & 67.68 & 68.65\\ 
        \rowcolor{LGray}
        Ours & ResNet50 & \textbf{98.01} & 87.08 & 74.75 \\ 
        \rowcolor{LGray}
        Ours & ViT-S/16 & 97.93 & \textbf{87.66}  & \textbf{78.00} \\
        \bottomrule
    \end{tabularx}
    \label{tab:obj3}
\end{table*}
\begin{figure*}[!t]
    \vspace{-1em}
    \centering
    \bgroup
    \def\arraystretch{1.5}
        \begin{tabular}{ccccccc}
    \includegraphics[width=0.145\textwidth,height=0.145\textwidth]{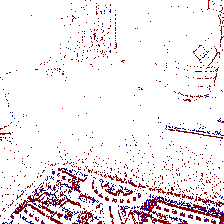} & 
       \includegraphics[width=0.145\textwidth,height=0.145\textwidth]{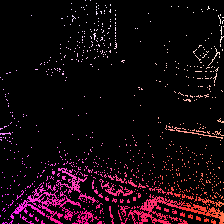} &
       \includegraphics[width=0.145\textwidth,height=0.145\textwidth]{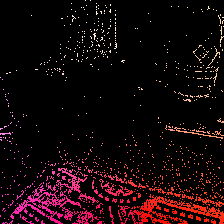} &       \includegraphics[width=0.145\textwidth,height=0.145\textwidth]{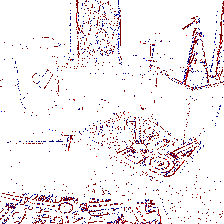} & 
       \includegraphics[width=0.145\textwidth,height=0.145\textwidth]{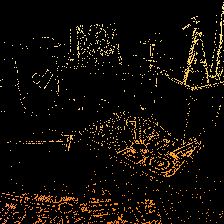} &
       \includegraphics[width=0.145\textwidth,height=0.145\textwidth]{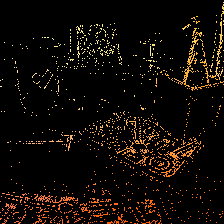} \\
       \includegraphics[width=0.145\textwidth,height=0.145\textwidth]{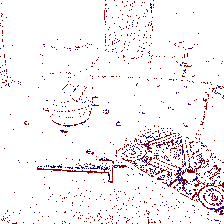} & 
       \includegraphics[width=0.145\textwidth,height=0.145\textwidth]{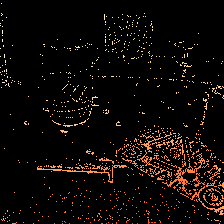} &
       \includegraphics[width=0.145\textwidth,height=0.145\textwidth]{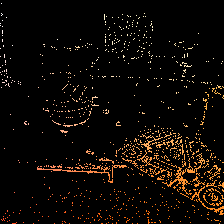} &       \includegraphics[width=0.145\textwidth,height=0.145\textwidth]{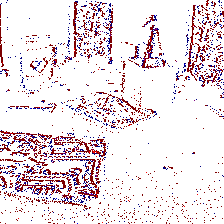} & 
       \includegraphics[width=0.145\textwidth,height=0.145\textwidth]{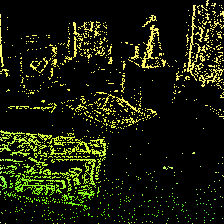} &
       \includegraphics[width=0.145\textwidth,height=0.145\textwidth]{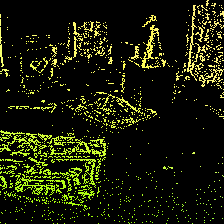} \\
       (a) & (b) & (c) &  (d) & (e) & (f) \\
    \end{tabular}
    \egroup
    \vspace{-1em}
    \caption{\it \small 
    Optical flow prediction examples of our method on the MVSEC dataset \cite{mvsec}. (a)/(d) are event images, where red and blue indicate positive and negative events.
    (b)/(e) are ground-truth optical flows.
    (c)/(f) are our predicted optical flows. }
    \label{fig:flow_mainpaper}
\end{figure*}

\vspace{-4mm}
\paragraph{Results on other small-scale datasets.}
The comparisons {on N-cars \cite{ncars}, N-Caltech101 \cite{ncaltech}, and CIFAR-10-DVS \cite{CIFAR-10-DVS} datasets} are given in \cref{tab:obj3}.
Note that the N-Caltech101 and CIFAR-10-DVS have not provided training and testing splits. We therefore randomly split them for generating training and testing datasets (please refer to the supplementary materials). 
Our pre-trained model with a ViT-S/16 backbone outperforms all other methods, with 97.93\%, 87.66\%, and 78.00\% top-1 accuracy on N-Cars, N-Caltech101, and CIFAR-10-DVS datasets, respectively.

\subsection{Optical Flow Estimation}
\label{sec::flow_estimation}
We show our optical flow estimation performance on the MVSEC dataset \cite{mvsec}.  Following \cite{mae,beit}, we simply append a decoder network to pre-trained networks to estimate the optical flow. Please refer to the supplementary material 
for architecture details and train-test splitting. The comparisons on the  `indoor\_flying1', `indoor\_flying2', and `indoor\_flying3' scenes are given in \cref{tab:flo}. 

Compared with other methods, our method with a ViT-S/16 backbone has the lowest AEEs and outlier ratios, showing the effectiveness of our pre-trained model for the optical flow estimation task. We show optical flow prediction examples of our method in \cref{fig:flow_mainpaper}.

\begin{table*}[!t]
    \centering
    \caption{\it  \small
    Comparison of optical flow estimation on the MVSEC dataset \cite{mvsec}. 
    We use average end-point error (AEE) and percentage of outliers (\%) for evaluation. Similar to the KITTI benchmark \cite{kitti}, the outlier measures the percentage of pixels that has end-point error larger than three units and  5\% of the ground truth optical flow.
    }
    \small
    \vspace{-1em}
    \begin{tabularx}{\linewidth}{lYYYcYYcYY}
        \toprule
        \multicolumn{1}{c}{\multirow{2}{*}{Method}} & \multirow{2}{*}{Backbone} & \multicolumn{2}{c}{\textit{indoor\_flying1}} && \multicolumn{2}{c}{\textit{indoor\_flying2}} && \multicolumn{2}{c}{\textit{indoor\_flying3}} \\
        \cmidrule{3-4} \cmidrule{6-7} \cmidrule{9-10}
        & & AEE & Outlier && AEE & Outlier && AEE & Outlier \\
        \midrule
        \multicolumn{9}{l}{\textit{The best performance in the literature. }}   \\ 
        EST~\cite{est} & - & 1.24 & 5.09 && 2.05 & 19.90 && 1.71 & 11.67\\
        DCEIFlow~\cite{dceiflow} & - & 0.75 & 0.60 && 1.39 & 8.01 && 1.13 & 5.29 \\
        \midrule
        \multicolumn{10}{l}{\textit{\makecell[l]{Training from scratch, \ie, random weight initialization.}}} \\
        ViT \cite{vit} & ViT-S/16 & 0.68 & 0.13 && 1.38 & 7.58 && 1.08 & 3.76\\
        ViT \cite{vit} & ViT-B/16 & 0.64 & 0.19 && 1.36 & 7.21 && 1.05 & 3.86\\
        ResNet \cite{resnet} & ResNet50 & 0.73 & 0.66 && 1.55 & 9.81 && 1.23 & 5.77\\
        \midrule
        \multicolumn{10}{l}{\textit{\makecell[l]{Transfer learning of supervised pre-training methods, \ie, initial weights learned in a supervised manner.}}} \\
        ViT \cite{vit} & ViT-S/16 & 0.88 & 3.06 && 1.79 & 16.63 && 1.49 & 8.66 \\
        ViT \cite{vit} & ViT-B/16 & 0.65 & 0.45 && 1.34 & 7.65 && 1.11 & 4.96\\
        ResNet \cite{resnet} & ResNet50  & \textbf{0.60} & 0.23 && 1.37 & 8.76 && 1.15 & 5.34 \\
        \midrule 
        \multicolumn{10}{l}{\textit{\makecell[l]{Transfer learning of self-supervised pre-training methods, \ie, initial weights learned in a self-supervised manner.}}} \\
        SimCLR \cite{simclr} & ResNet50 & 0.65 & 0.49 && 1.45 & 9.33 && 1.19 & 5.51  \\
        MoCo-v2 \cite{mocov2} & ResNet50 & 0.61 & 0.46 && 1.36 & 8.68 && 1.13 & 5.20\\
        MoCo-v3 \cite{mocov3} & ViT-S/16 & 0.66 & 0.35 && 1.41 & 8.23 && 1.17 & 5.10\\
        BeiT \cite{beit} & ViT-B/16 & 0.64 & 0.29 && 1.32 & 7.34 && 1.07 & 4.32 \\
        iBoT \cite{ibot} & ViT-S/16 & 0.80 & 0.81 && 1.47 & 8.77 && 1.16 & 5.43 \\
        MAE \cite{mae} & ViT-B/16 & 0.61 & 0.17 && 1.29 & 6.95 && 1.11 & 4.64 \\
        \rowcolor{LGray}
        Ours & ResNet50 & \textbf{0.60} & 0.35 && 1.35 & 8.57  && 1.12 & 5.26 \\
        \rowcolor{LGray}
        Ours & ViT-S/16 & 0.61 & \textbf{0.05} && \textbf{1.26} & \textbf{6.69}  && \textbf{1.00} & \textbf{3.11}\\
        \bottomrule
     \end{tabularx}
    \label{tab:flo}
\end{table*}

\subsection{Semantic Segmentation}\label{sec::semantic_segmentation}
We show our semantic segmentation performance on the DDD17 \cite{ddd17,evsegnet} and DSEC datasets \cite{dsec,ess}. {
Following \cite{beit}, we simply append a decoder network to pre-trained networks to estimate semantic labels, and use the mean interaction over union (mIoU) metric to evaluate methods. The comparisons are given in \cref{tab:sem}.}

\begin{figure*}[!t]
    \vspace{-1em}
    \centering
    \bgroup
    \def\arraystretch{1.5}
        \begin{tabular}{ccccccc}
    \includegraphics[width=0.145\textwidth,height=0.145\textwidth]{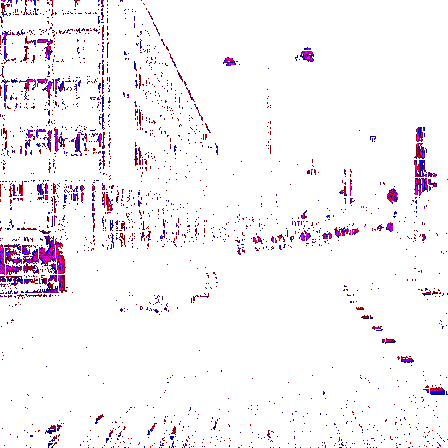} & 
       \includegraphics[width=0.145\textwidth,height=0.145\textwidth]{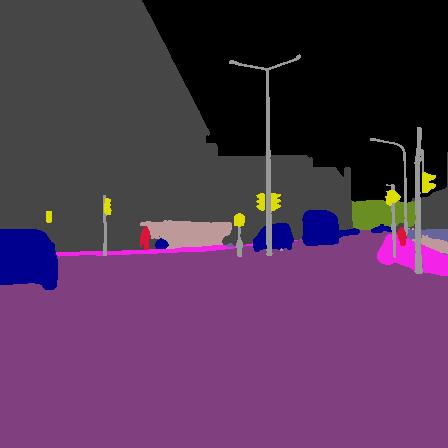} &
       \includegraphics[width=0.145\textwidth,height=0.145\textwidth]{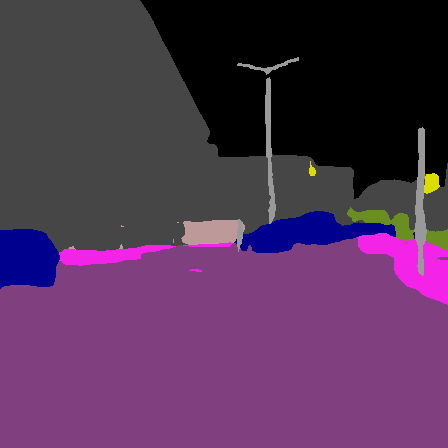} &       \includegraphics[width=0.145\textwidth,height=0.145\textwidth]{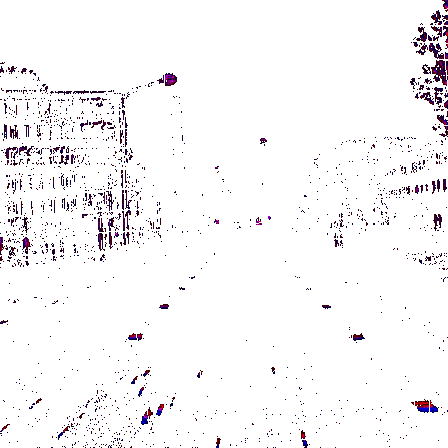} & 
       \includegraphics[width=0.145\textwidth,height=0.145\textwidth]{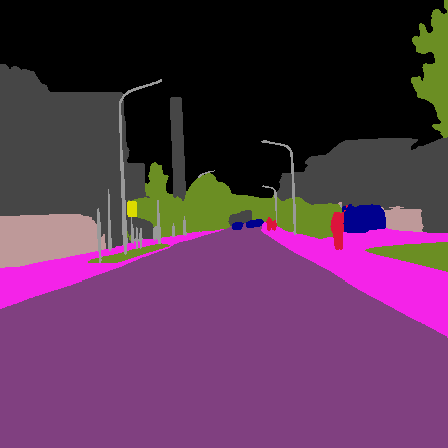} &
       \includegraphics[width=0.145\textwidth,height=0.145\textwidth]{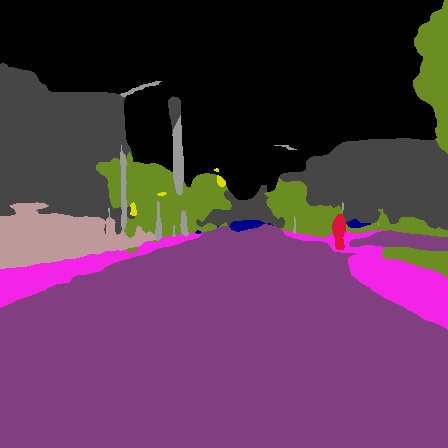} \\

       \includegraphics[width=0.145\textwidth,height=0.145\textwidth]{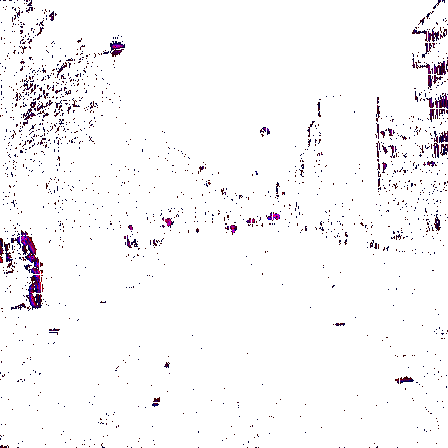} & 
       \includegraphics[width=0.145\textwidth,height=0.145\textwidth]{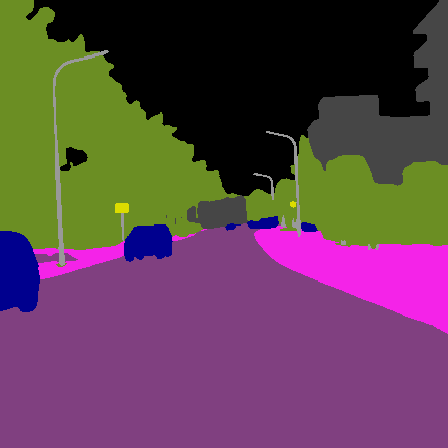} &
       \includegraphics[width=0.145\textwidth,height=0.145\textwidth]{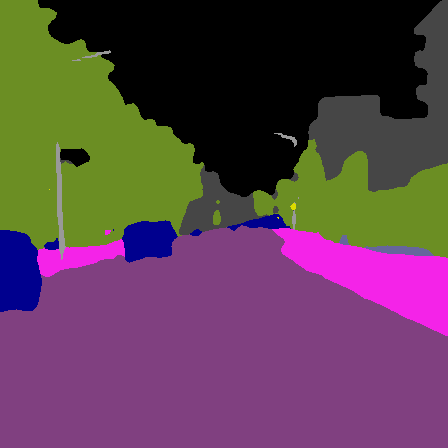} &       \includegraphics[width=0.145\textwidth,height=0.145\textwidth]{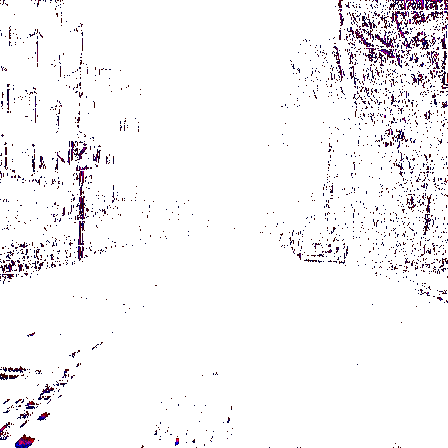} & 
       \includegraphics[width=0.145\textwidth,height=0.145\textwidth]{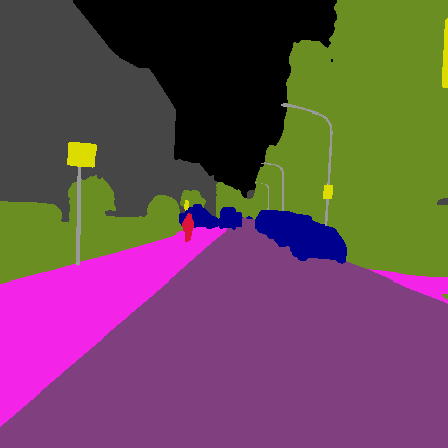} &
       \includegraphics[width=0.145\textwidth,height=0.145\textwidth]{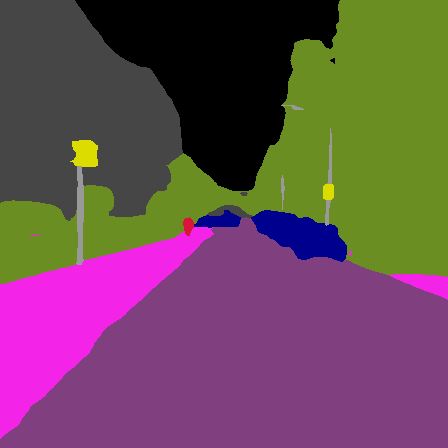} \\
       (a) & (b) & (c) &  (d) & (e) & (f) \\
    \end{tabular}
    \egroup
    \vspace{-1em}
    \caption{\it \small 
    Semantic segmentation prediction examples of our method on the DSEC dataset \cite{mvsec}. (a)/(d) are event images, where red and blue indicate positive and negative events.
    (b)/(e) are ground-truth segmentation images, and pixel colors denote semantic classes.
    (c)/(f) are our predicted segmentation images.}
    \label{fig:seg_mainpaper}
\end{figure*}

The performance of our method with a ResNet50 backbone is comparable with respect to the state-of-the-art method ESS~\cite{ess}, which uses additional RGB images and their semantic labels for training. For methods only using event data and semantic labels for training, our method outperforms the state-of-the-art method EV-SegNet~\cite{evsegnet}. Please refer to \cref{fig:seg_mainpaper} for our semantic segmentation examples.

\begin{table}[!t]
    \centering
    \caption{\small \it Comparison of semantic segmentation on the DDD17 \cite{ddd17,evsegnet} and DSEC datasets \cite{dsec,ess}. Following \cite{beit}, we use the mean interaction over union (mIoU (\%)) for comparison. Our method and EV-SegNet~\cite{evsegnet} only use  event data and corresponding semantic labels for training. ESS~\cite{ess} uses additional RGB images and semantic labels in the training stage.
    }
    \small
    \vspace{-1em}
    \begin{tabularx}{\linewidth}{lXXX}
        \toprule
         \multicolumn{1}{c}{Method} & Backbone & DDD17 & DSEC \\
        \midrule
        \multicolumn{4}{l}{\textit{\makecell[l]{The best performance in the literature.}}} \\
        EV-SegNet \cite{evsegnet}  &  -  & 54.81 & 51.76\\
        ESS \cite{ess}   &  -  & \textbf{61.37} & 53.29\\
        \midrule
        \multicolumn{4}{l}{\textit{\makecell[l]{Training from scratch.}}} \\
        ViT \cite{vit} & ViT-S/16 & 48.76 & 40.53\\
        ViT \cite{vit} & ViT-B/16 & 43.89 & 38.24\\
        ResNet & ResNet50 & 56.96 & 57.60\\
        \midrule
        \multicolumn{4}{l}{\textit{\makecell[l]{Transfer learning of supervised pre-training methods.}}} \\
        ViT \cite{vit} & ViT-S/16 & 54.12 & 42.92\\ 
        ViT \cite{vit} & ViT-B/16 & 54.06 & 45.55\\
        ResNet & ResNet50 & 59.25 & 58.50\\
        \midrule 
        \multicolumn{4}{l}{\textit{\makecell[l]{Transfer learning of self-supervised pre-training methods. }}} \\
        SimCLR \cite{simclr} & ResNet50 & 57.22 & 59.06\\
        MoCo-v2 \cite{mocov2} & ResNet50  & 58.28 & 59.09\\
        MoCo-v3 \cite{mocov3} & ViT-S/16  & 53.65 & 49.21\\
        BeiT \cite{beit} & ViT-B/16 & 52.39 & 46.52\\
        IBoT \cite{ibot} & ViT-S/16 & 49.94 & 42.53\\
        MAE \cite{mae} & ViT-B/16 & 52.36 & 47.56\\
        \rowcolor{LGray}
        Ours & ViT-S/16 & 54.66 & 47.91\\
        \rowcolor{LGray}
        Ours & ResNet50 & 59.15 & \textbf{59.16}\\
        \bottomrule
    \end{tabularx}
    \label{tab:sem}
\end{table}

\begin{figure*}[!t]
    \centering
    \bgroup
    \def\arraystretch{1.5}
    \begin{tabular}{ccccccc}
     \includegraphics[width=0.145\textwidth,height=0.145\textwidth]{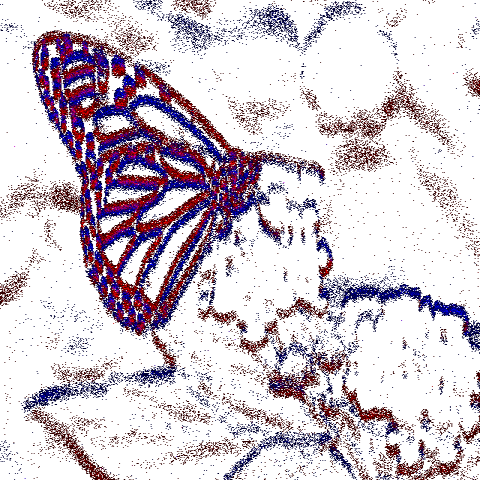} 
     & \includegraphics[width=0.145\textwidth,height=0.145\textwidth]{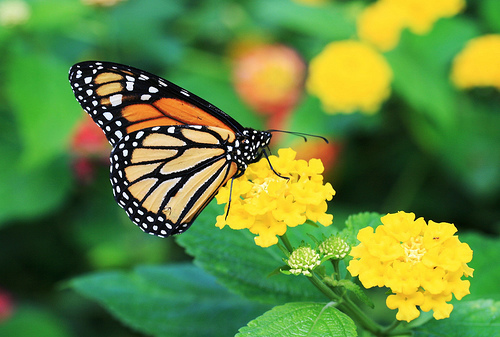} 
     & \includegraphics[width=0.145\textwidth,height=0.145\textwidth]{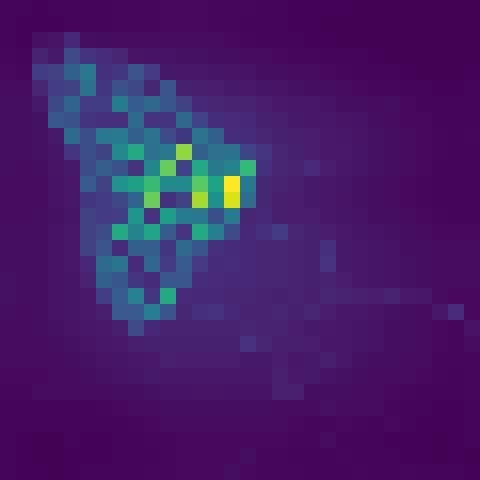} &       \includegraphics[width=0.145\textwidth,height=0.145\textwidth]{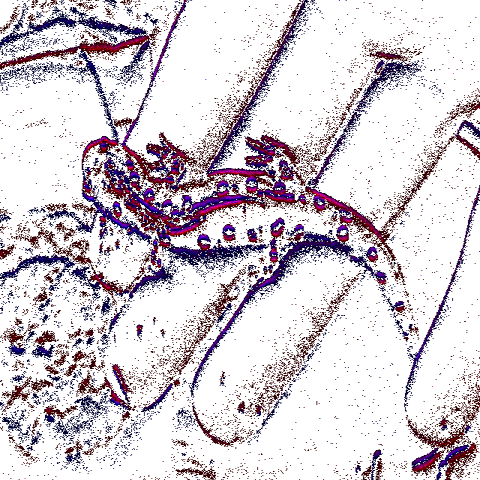} & 
       \includegraphics[width=0.145\textwidth,height=0.145\textwidth]{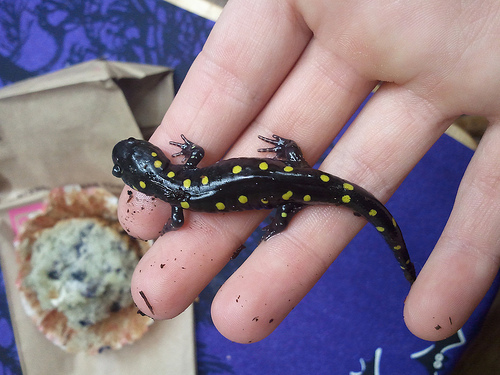}
       & 
       \includegraphics[width=0.145\textwidth,height=0.145\textwidth]{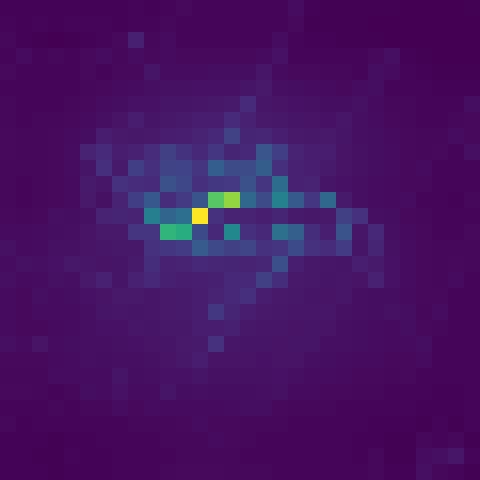} \\
       \includegraphics[width=0.145\textwidth,height=0.145\textwidth]{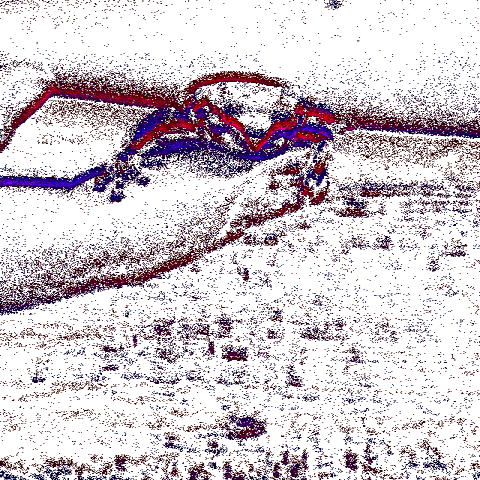} & 
       \includegraphics[width=0.145\textwidth,height=0.145\textwidth]{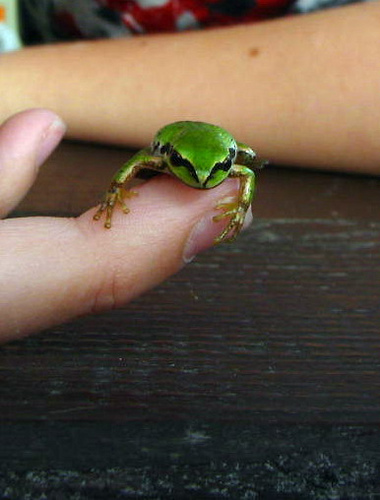} &
       \includegraphics[width=0.145\textwidth,height=0.145\textwidth]{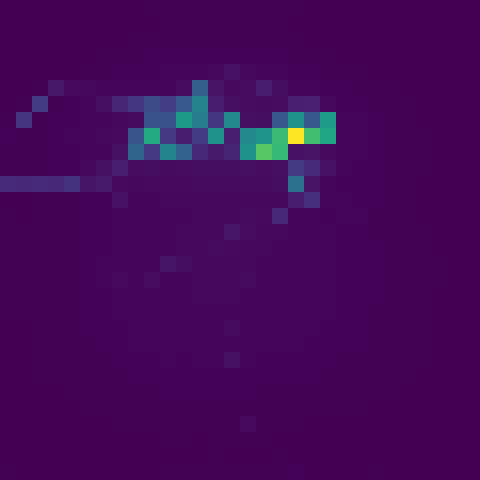} & \includegraphics[width=0.145\textwidth,height=0.145\textwidth]{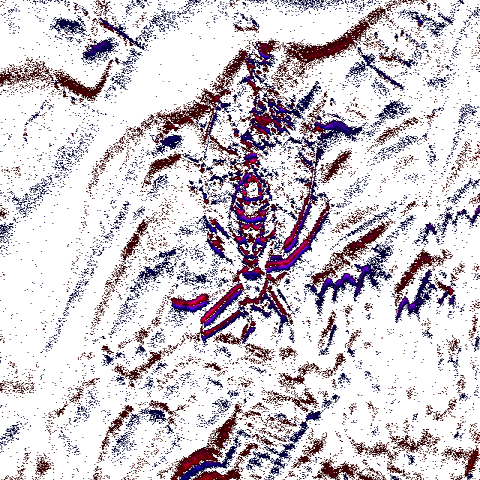} & 
       \includegraphics[width=0.145\textwidth,height=0.145\textwidth]{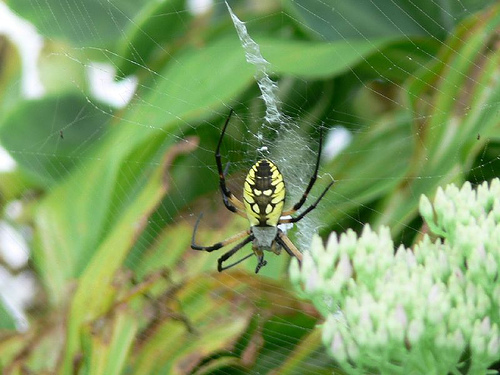} &
       \includegraphics[width=0.145\textwidth,height=0.145\textwidth]{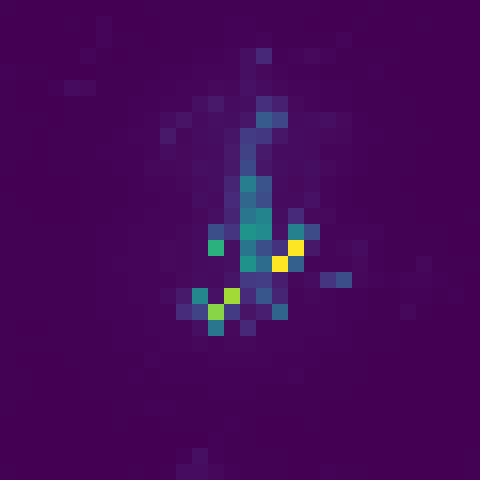} \\
       (a) & (b) & (c) &  (d) & (e) & (f) \\
    \end{tabular}
    \egroup
    \vspace{-1em}
    \caption{\it \small
    Attention maps of our pre-trained model (without any fine-tuning) on sample data from the N-ImageNet dataset \cite{nimagnet}.  
    (a)/(d) are event images.  Similarly, we use red and blue to indicate positive and negative events. (b)/(e) are corresponding natural RGB images used for visualization assistance. 
    (c)/(f) are our attention maps.
    }
    \label{fig:img}
\end{figure*}

\subsection{Discussion}

\paragraph{Analysis of attention maps.}  We visualize attention maps of our pre-trained model in \cref{fig:img}, where features from the last layer of our pre-trained model are used to compute the attention map. The results show that our pre-trained model successfully focuses on semantic meaningful objects on the noisy event images (\eg, spider on the second row of \cref{fig:img}). This strong pattern discovery ability of our method potentially explains the effectiveness of our pre-trained model when transferring to diverse downstream tasks.

\vspace{-4mm}
\paragraph{Effectiveness of data augmentations.} Generating different views of the same data is one of the most important parts of the self-supervised learning framework. 
We compare two methods: i) previously commonly used methods \cite{mocov3,mae}, ii) our event data augmentations. By pre-training our method with the above two different augmentation methods for 100 epochs on the N-ImageNet, we obtain 46.1\% and 53.5\% top-1 accuracy with linear probing, respectively. 

\vspace{-4mm}
\paragraph{Pre-train MAE using event data?} 
Can we pre-train the state-of-the-art self-supervised method MAE~\cite{mae} using event data? To check the feasibility, we perform a binary search to find the best masking ratio and optimization schema for MAE. MAE with a ViT-B/16 backbone obtains top-1 accuracy at 55.45\%  after fine-tuning on the N-ImageNet dataset. In comparison, our method with a ViT-S/16 backbone achieves top-1 accuracy at 64.83\%.

\begin{table}[!t]
    \centering
    \caption{\small \it Scaling the number of parameters of $f_{\mathsf{e}}$ and $f_{\mathsf{m}}$.}
    \vspace{-1em}
    \small
    \begin{tabularx}{\linewidth}{lYYcYY}
        \toprule
         \multirow{2}{*}{Backbone} & \multicolumn{2}{c}{Linear Probing} & ~ & \multicolumn{2}{c}{Fine-tuning} \\
         \cmidrule{2-3} \cmidrule{5-6}
         & acc@1 & acc@5 && acc@1 & acc@5 \\
         \midrule
         ViT-S/16 & 59.90 & 82.26 && 64.84 & 86.30\\
         ViT-B/16 & 61.75 & 82.53 && 68.31 & 88.02\\
         ViT-L/16 & 64.53 & 84.90 && 71.05 & 89.86\\
         \bottomrule
    \end{tabularx}
    \label{tab:arc}
\end{table}

\vspace{-4mm}
\paragraph{Model Scalability.} We scale our backbone ($f_{\mathsf{e}}$ and $f_{\mathsf{m}}$) from ViT-S/16 to ViT-L/16. The results are given in \cref{tab:arc}. The accuracy of our method improves with respect to the increasing number of model parameters of ViT. 

\section{Conclusion} 
In this paper, we have trained a neural network for processing event camera data, as a self-supervised learning framework. The method contains three key components: a family of event data augmentations, a conditional masking strategy, and a contrastive learning approach. Our key insight is enforcing the similarity of embeddings between matching event images and between paired event and RGB images to pre-train our model. Extensive experiments on downstream tasks (\ie, object recognition, optical flow estimation, and semantic segmentation) demonstrate the superiority of our method over past methods. 

\vspace{-4mm}
\paragraph{Broader impacts.} There are diverse potential extensions of our method. For example, our model is promising to achieve zero-shot or few-shot learning by using existing vision-language methods from the RGB image domain, due to event data and RGB images being aligned in the same feature space. We hope this paper will inspire future work.

{\small
\bibliographystyle{ieee_fullname}
\bibliography{egbib}

\begin{thebibliography}{10}\itemsep=-1pt

\bibitem{evsegnet}
I{\~{n}}igo Alonso and Ana~C. Murillo.
\newblock Ev-segnet: Semantic segmentation for event-based cameras.
\newblock In {\em {IEEE} Conference on Computer Vision and Pattern Recognition
  Workshops, {CVPR} Workshops 2019, Long Beach, CA, USA, June 16-20, 2019},
  pages 1624--1633. Computer Vision Foundation / {IEEE}, 2019.

\bibitem{beit}
Hangbo Bao, Li Dong, Songhao Piao, and Furu Wei.
\newblock Beit: {BERT} pre-training of image transformers.
\newblock In {\em The Tenth International Conference on Learning
  Representations, {ICLR} 2022, Virtual Event, April 25-29, 2022}.
  OpenReview.net, 2022.

\bibitem{ddd17}
Jonathan Binas, Daniel Neil, Shih{-}Chii Liu, and Tobi Delbr{\"{u}}ck.
\newblock {DDD17:} end-to-end {DAVIS} driving dataset.
\newblock {\em CoRR}, abs/1711.01458, 2017.

\bibitem{swav}
Mathilde Caron, Ishan Misra, Julien Mairal, Priya Goyal, Piotr Bojanowski, and
  Armand Joulin.
\newblock Unsupervised learning of visual features by contrasting cluster
  assignments.
\newblock In Hugo Larochelle, Marc'Aurelio Ranzato, Raia Hadsell,
  Maria{-}Florina Balcan, and Hsuan{-}Tien Lin, editors, {\em Advances in
  Neural Information Processing Systems 33: Annual Conference on Neural
  Information Processing Systems 2020, NeurIPS 2020, December 6-12, 2020,
  virtual}, 2020.

\bibitem{dino}
Mathilde Caron, Hugo Touvron, Ishan Misra, Herv{\'{e}} J{\'{e}}gou, Julien
  Mairal, Piotr Bojanowski, and Armand Joulin.
\newblock Emerging properties in self-supervised vision transformers.
\newblock In {\em 2021 {IEEE/CVF} International Conference on Computer Vision,
  {ICCV} 2021, Montreal, QC, Canada, October 10-17, 2021}, pages 9630--9640.
  {IEEE}, 2021.

\bibitem{igpt}
Mark Chen, Alec Radford, Rewon Child, Jeffrey Wu, Heewoo Jun, David Luan, and
  Ilya Sutskever.
\newblock Generative pretraining from pixels.
\newblock In {\em Proceedings of the 37th International Conference on Machine
  Learning, {ICML} 2020, 13-18 July 2020, Virtual Event}, volume 119 of {\em
  Proceedings of Machine Learning Research}, pages 1691--1703. {PMLR}, 2020.

\bibitem{simclr}
Ting Chen, Simon Kornblith, Mohammad Norouzi, and Geoffrey~E. Hinton.
\newblock A simple framework for contrastive learning of visual
  representations.
\newblock In {\em Proceedings of the 37th International Conference on Machine
  Learning, {ICML} 2020, 13-18 July 2020, Virtual Event}, volume 119 of {\em
  Proceedings of Machine Learning Research}, pages 1597--1607. {PMLR}, 2020.

\bibitem{mocov2}
Xinlei Chen, Haoqi Fan, Ross Girshick, and Kaiming He.
\newblock Improved baselines with momentum contrastive learning.
\newblock {\em arXiv preprint arXiv:2003.04297}, 2020.

\bibitem{simsiam}
Xinlei Chen and Kaiming He.
\newblock Exploring simple siamese representation learning.
\newblock In {\em {IEEE} Conference on Computer Vision and Pattern Recognition,
  {CVPR} 2021, virtual, June 19-25, 2021}, pages 15750--15758. Computer Vision
  Foundation / {IEEE}, 2021.

\bibitem{mocov3}
Xinlei Chen*, Saining Xie*, and Kaiming He.
\newblock An empirical study of training self-supervised vision transformers.
\newblock {\em arXiv preprint arXiv:2104.02057}, 2021.

\bibitem{CIFAR-10-DVS}
Wensheng Cheng, Hao Luo, Wen Yang, Lei Yu, and Wei Li.
\newblock Structure-aware network for lane marker extraction with dynamic
  vision sensor.
\newblock {\em CoRR}, abs/2008.06204, 2020.

\bibitem{imagenet}
Jia Deng, Wei Dong, Richard Socher, Li{-}Jia Li, Kai Li, and Li Fei{-}Fei.
\newblock Imagenet: {A} large-scale hierarchical image database.
\newblock In {\em 2009 {IEEE} Computer Society Conference on Computer Vision
  and Pattern Recognition {(CVPR} 2009), 20-25 June 2009, Miami, Florida,
  {USA}}, pages 248--255. {IEEE} Computer Society, 2009.

\bibitem{bert}
Jacob Devlin, Ming{-}Wei Chang, Kenton Lee, and Kristina Toutanova.
\newblock {BERT:} pre-training of deep bidirectional transformers for language
  understanding.
\newblock In Jill Burstein, Christy Doran, and Thamar Solorio, editors, {\em
  Proceedings of the 2019 Conference of the North American Chapter of the
  Association for Computational Linguistics: Human Language Technologies,
  {NAACL-HLT} 2019, Minneapolis, MN, USA, June 2-7, 2019, Volume 1 (Long and
  Short Papers)}, pages 4171--4186. Association for Computational Linguistics,
  2019.

\bibitem{vit}
Alexey Dosovitskiy, Lucas Beyer, Alexander Kolesnikov, Dirk Weissenborn,
  Xiaohua Zhai, Thomas Unterthiner, Mostafa Dehghani, Matthias Minderer, Georg
  Heigold, Sylvain Gelly, Jakob Uszkoreit, and Neil Houlsby.
\newblock An image is worth 16x16 words: Transformers for image recognition at
  scale.
\newblock In {\em 9th International Conference on Learning Representations,
  {ICLR} 2021, Virtual Event, Austria, May 3-7, 2021}. OpenReview.net, 2021.

\bibitem{vqgan}
Patrick Esser, Robin Rombach, and Bj{\"{o}}rn Ommer.
\newblock Taming transformers for high-resolution image synthesis.
\newblock In {\em {IEEE} Conference on Computer Vision and Pattern Recognition,
  {CVPR} 2021, virtual, June 19-25, 2021}, pages 12873--12883. Computer Vision
  Foundation / {IEEE}, 2021.

\bibitem{est}
Daniel Gehrig, Antonio Loquercio, Konstantinos~G. Derpanis, and Davide
  Scaramuzza.
\newblock End-to-end learning of representations for asynchronous event-based
  data.
\newblock In {\em 2019 {IEEE/CVF} International Conference on Computer Vision,
  {ICCV} 2019, Seoul, Korea (South), October 27 - November 2, 2019}, pages
  5632--5642. {IEEE}, 2019.

\bibitem{dsec}
Mathias Gehrig, Willem Aarents, Daniel Gehrig, and Davide Scaramuzza.
\newblock {DSEC:} {A} stereo event camera dataset for driving scenarios.
\newblock {\em {IEEE} Robotics Autom. Lett.}, 6(3):4947--4954, 2021.

\bibitem{byol}
Jean{-}Bastien Grill, Florian Strub, Florent Altch{\'{e}}, Corentin Tallec,
  Pierre~H. Richemond, Elena Buchatskaya, Carl Doersch, Bernardo~{\'{A}}vila
  Pires, Zhaohan Guo, Mohammad~Gheshlaghi Azar, Bilal Piot, Koray Kavukcuoglu,
  R{\'{e}}mi Munos, and Michal Valko.
\newblock Bootstrap your own latent - {A} new approach to self-supervised
  learning.
\newblock In Hugo Larochelle, Marc'Aurelio Ranzato, Raia Hadsell,
  Maria{-}Florina Balcan, and Hsuan{-}Tien Lin, editors, {\em Advances in
  Neural Information Processing Systems 33: Annual Conference on Neural
  Information Processing Systems 2020, NeurIPS 2020, December 6-12, 2020,
  virtual}, 2020.

\bibitem{mae}
Kaiming He, Xinlei Chen, Saining Xie, Yanghao Li, Piotr Doll{\'{a}}r, and
  Ross~B. Girshick.
\newblock Masked autoencoders are scalable vision learners.
\newblock In {\em {IEEE/CVF} Conference on Computer Vision and Pattern
  Recognition, {CVPR} 2022, New Orleans, LA, USA, June 18-24, 2022}, pages
  15979--15988. {IEEE}, 2022.

\bibitem{mocov1}
Kaiming He, Haoqi Fan, Yuxin Wu, Saining Xie, and Ross~B. Girshick.
\newblock Momentum contrast for unsupervised visual representation learning.
\newblock In {\em 2020 {IEEE/CVF} Conference on Computer Vision and Pattern
  Recognition, {CVPR} 2020, Seattle, WA, USA, June 13-19, 2020}, pages
  9726--9735. Computer Vision Foundation / {IEEE}, 2020.

\bibitem{resnet}
Kaiming He, Xiangyu Zhang, Shaoqing Ren, and Jian Sun.
\newblock Deep residual learning for image recognition.
\newblock {\em CoRR}, abs/1512.03385, 2015.

\bibitem{nimagnet}
Junho Kim, Jaehyeok Bae, Gangin Park, Dongsu Zhang, and Young~Min Kim.
\newblock N-imagenet: Towards robust, fine-grained object recognition with
  event cameras.
\newblock In {\em Proceedings of the IEEE/CVF International Conference on
  Computer Vision (ICCV)}, pages 2146--2156, October 2021.

\bibitem{eventhistorgram}
Ana~I. Maqueda, Antonio Loquercio, Guillermo Gallego, Narciso Garc{\'{\i}}a,
  and Davide Scaramuzza.
\newblock Event-based vision meets deep learning on steering prediction for
  self-driving cars.
\newblock In {\em 2018 {IEEE} Conference on Computer Vision and Pattern
  Recognition, {CVPR} 2018, Salt Lake City, UT, USA, June 18-22, 2018}, pages
  5419--5427. Computer Vision Foundation / {IEEE} Computer Society, 2018.

\bibitem{kitti}
M. Menze, Christian Heipke, and Andreas Geiger.
\newblock Joint 3d estimation of vehicles and scene flow.
\newblock {\em ISPRS Annals of Photogrammetry, Remote Sensing and Spatial
  Information Sciences}, II-3/W5:427--434, 08 2015.

\bibitem{ed1}
Nico Messikommer, Daniel Gehrig, Mathias Gehrig, and Davide Scaramuzza.
\newblock Bridging the gap between events and frames through unsupervised
  domain adaptation.
\newblock {\em {IEEE} Robotics Autom. Lett.}, 7(2):3515--3522, 2022.

\bibitem{rc2}
Nico Messikommer, Daniel Gehrig, Antonio Loquercio, and Davide Scaramuzza.
\newblock Event-based asynchronous sparse convolutional networks.
\newblock In Andrea Vedaldi, Horst Bischof, Thomas Brox, and Jan{-}Michael
  Frahm, editors, {\em Computer Vision - {ECCV} 2020 - 16th European
  Conference, Glasgow, UK, August 23-28, 2020, Proceedings, Part {VIII}},
  volume 12353 of {\em Lecture Notes in Computer Science}, pages 415--431.
  Springer, 2020.

\bibitem{od3}
Anindya Mondal, Shashant R, Jhony~H. Giraldo, Thierry Bouwmans, and Ananda~S.
  Chowdhury.
\newblock Moving object detection for event-based vision using graph spectral
  clustering.
\newblock In {\em {IEEE/CVF} International Conference on Computer Vision
  Workshops, {ICCVW} 2021, Montreal, BC, Canada, October 11-17, 2021}, pages
  876--884. {IEEE}, 2021.

\bibitem{ncaltech}
Garrick Orchard, Ajinkya Jayawant, Gregory Cohen, and Nitish~V. Thakor.
\newblock Converting static image datasets to spiking neuromorphic datasets
  using saccades.
\newblock {\em CoRR}, abs/1507.07629, 2015.

\bibitem{pytorch}
Adam Paszke, Sam Gross, Francisco Massa, Adam Lerer, James Bradbury, Gregory
  Chanan, Trevor Killeen, Zeming Lin, Natalia Gimelshein, Luca Antiga, Alban
  Desmaison, Andreas K{\"{o}}pf, Edward~Z. Yang, Zachary DeVito, Martin Raison,
  Alykhan Tejani, Sasank Chilamkurthy, Benoit Steiner, Lu Fang, Junjie Bai, and
  Soumith Chintala.
\newblock Pytorch: An imperative style, high-performance deep learning library.
\newblock In Hanna~M. Wallach, Hugo Larochelle, Alina Beygelzimer, Florence
  d'Alch{\'{e}}{-}Buc, Emily~B. Fox, and Roman Garnett, editors, {\em Advances
  in Neural Information Processing Systems 32: Annual Conference on Neural
  Information Processing Systems 2019, NeurIPS 2019, December 8-14, 2019,
  Vancouver, BC, Canada}, pages 8024--8035, 2019.

\bibitem{od1}
Etienne Perot, Pierre de Tournemire, Davide Nitti, Jonathan Masci, and Amos
  Sironi.
\newblock Learning to detect objects with a 1 megapixel event camera.
\newblock In Hugo Larochelle, Marc'Aurelio Ranzato, Raia Hadsell,
  Maria{-}Florina Balcan, and Hsuan{-}Tien Lin, editors, {\em Advances in
  Neural Information Processing Systems 33: Annual Conference on Neural
  Information Processing Systems 2020, NeurIPS 2020, December 6-12, 2020,
  virtual}, 2020.

\bibitem{gpt}
Alec Radford, Jeff Wu, Rewon Child, David Luan, Dario Amodei, and Ilya
  Sutskever.
\newblock Language models are unsupervised multitask learners.
\newblock {\em OpenAI}, 2019.

\bibitem{dvae}
Jason~Tyler Rolfe.
\newblock Discrete variational autoencoders.
\newblock In {\em 5th International Conference on Learning Representations,
  {ICLR} 2017, Toulon, France, April 24-26, 2017, Conference Track
  Proceedings}. OpenReview.net, 2017.

\bibitem{rc1}
Cedric Scheerlinck, Nick Barnes, and Robert~E. Mahony.
\newblock Continuous-time intensity estimation using event cameras.
\newblock In C.~V. Jawahar, Hongdong Li, Greg Mori, and Konrad Schindler,
  editors, {\em Computer Vision - {ACCV} 2018 - 14th Asian Conference on
  Computer Vision, Perth, Australia, December 2-6, 2018, Revised Selected
  Papers, Part {V}}, volume 11365 of {\em Lecture Notes in Computer Science},
  pages 308--324. Springer, 2018.

\bibitem{eventtoimage}
Cedric Scheerlinck, Henri Rebecq, Daniel Gehrig, Nick Barnes, Robert~E. Mahony,
  and Davide Scaramuzza.
\newblock Fast image reconstruction with an event camera.
\newblock In {\em {IEEE} Winter Conference on Applications of Computer Vision,
  {WACV} 2020, Snowmass Village, CO, USA, March 1-5, 2020}, pages 156--163.
  {IEEE}, 2020.

\bibitem{ncars}
Amos Sironi, Manuele Brambilla, Nicolas Bourdis, Xavier Lagorce, and Ryad
  Benosman.
\newblock {HATS:} histograms of averaged time surfaces for robust event-based
  object classification.
\newblock In {\em 2018 {IEEE} Conference on Computer Vision and Pattern
  Recognition, {CVPR} 2018, Salt Lake City, UT, USA, June 18-22, 2018}, pages
  1731--1740. Computer Vision Foundation / {IEEE} Computer Society, 2018.

\bibitem{ess}
Zhaoning Sun, Nico Messikommer, Daniel Gehrig, and Davide Scaramuzza.
\newblock {ESS:} learning event-based semantic segmentation from still images.
\newblock In Shai Avidan, Gabriel~J. Brostow, Moustapha Ciss{\'{e}},
  Giovanni~Maria Farinella, and Tal Hassner, editors, {\em Computer Vision -
  {ECCV} 2022 - 17th European Conference, Tel Aviv, Israel, October 23-27,
  2022, Proceedings, Part {XXXIV}}, volume 13694 of {\em Lecture Notes in
  Computer Science}, pages 341--357. Springer, 2022.

\bibitem{infonce}
A{\"{a}}ron van~den Oord, Yazhe Li, and Oriol Vinyals.
\newblock Representation learning with contrastive predictive coding.
\newblock {\em CoRR}, abs/1807.03748, 2018.

\bibitem{dceiflow}
Zhexiong Wan, Yuchao Dai, and Yuxin Mao.
\newblock Learning dense and continuous optical flow from an event camera.
\newblock {\em {IEEE} Trans. Image Process.}, 31:7237--7251, 2022.

\bibitem{rc3}
Lin Wang, S.~Mohammad~Mostafavi I., Yo{-}Sung Ho, and Kuk{-}Jin Yoon.
\newblock Event-based high dynamic range image and very high frame rate video
  generation using conditional generative adversarial networks.
\newblock In {\em {IEEE} Conference on Computer Vision and Pattern Recognition,
  {CVPR} 2019, Long Beach, CA, USA, June 16-20, 2019}, pages 10081--10090.
  Computer Vision Foundation / {IEEE}, 2019.

\bibitem{slam}
David Weikersdorfer, David~B. Adrian, Daniel Cremers, and J{\"{o}}rg Conradt.
\newblock Event-based 3d {SLAM} with a depth-augmented dynamic vision sensor.
\newblock In {\em 2014 {IEEE} International Conference on Robotics and
  Automation, {ICRA} 2014, Hong Kong, China, May 31 - June 7, 2014}, pages
  359--364. {IEEE}, 2014.

\bibitem{instancediscrmination}
Zhirong Wu, Yuanjun Xiong, Stella~X. Yu, and Dahua Lin.
\newblock Unsupervised feature learning via non-parametric instance
  discrimination.
\newblock In {\em 2018 {IEEE} Conference on Computer Vision and Pattern
  Recognition, {CVPR} 2018, Salt Lake City, UT, USA, June 18-22, 2018}, pages
  3733--3742. Computer Vision Foundation / {IEEE} Computer Society, 2018.

\bibitem{ibot}
Jinghao Zhou, Chen Wei, Huiyu Wang, Wei Shen, Cihang Xie, Alan~L. Yuille, and
  Tao Kong.
\newblock Image {BERT} pre-training with online tokenizer.
\newblock In {\em The Tenth International Conference on Learning
  Representations, {ICLR} 2022, Virtual Event, April 25-29, 2022}.
  OpenReview.net, 2022.

\bibitem{mvsec}
Alex~Zihao Zhu, Dinesh Thakur, Tolga {\"{O}}zaslan, Bernd Pfrommer, Vijay
  Kumar, and Kostas Daniilidis.
\newblock The multi vehicle stereo event camera dataset: An event camera
  dataset for 3d perception.
\newblock {\em CoRR}, abs/1801.10202, 2018.

\end{thebibliography}
}

\end{document}